\documentclass[10pt,journal,compsoc]{IEEEtran}
\ifCLASSOPTIONcompsoc
  \usepackage[nocompress]{cite}
\else
  \usepackage{cite}
\fi
\ifCLASSINFOpdf
\else
\fi
\usepackage{graphicx}
\usepackage{makecell}
\usepackage{xcolor}
\definecolor{mygreen}{rgb}{0, 0.6, 0}
\definecolor{accepted}{rgb}{0, 0.0, 0.6}
\definecolor{cut}{rgb}{1, 0.0, 0.0}

\usepackage{algorithmic}

\begin{document}
%
\title{Are you really looking at me? A Feature-Extraction Framework for Estimating Interpersonal Eye Gaze from Conventional Video}

%
%
%
%

\author{Minh~Tran,
        Taylan~Sen,
        Kurtis~Haut,
        Mohammad~Rafayet~Ali,~\IEEEmembership{Student Member,~IEEE}
        and~Mohammed~Ehsan~Hoque,~\IEEEmembership{Member,~IEEE}
\IEEEcompsocitemizethanks{\IEEEcompsocthanksitem Minh Tran and Kurtis Haut are with University of Rochester, Rochester, NY, 14627. \protect\\
E-mail: \{mtran14,khaut\}@u.rochester.edu
\IEEEcompsocthanksitem Taylan Sen, Mohammad Rafayet Ali and Mohammed Ehsan Hoque are with the Department
of Computer Science, University of Rochester, Rochester,
NY, 14627. \protect\\
E-mail: \{tsen,mali7,mehoque\}@cs.rochester.edu}
}

\IEEEtitleabstractindextext{%
\begin{abstract}

Despite a revolution in the pervasiveness of video cameras in our daily lives, one of the most meaningful forms of nonverbal affective communication, interpersonal eye gaze, i.e. eye gaze relative to a conversation partner, is not available from common video. We introduce the Interpersonal-Calibrating Eye-gaze Encoder (ICE), which automatically extracts interpersonal gaze from video recordings without specialized hardware and without prior knowledge of participant locations. Leveraging the intuition that individuals spend a large portion of a conversation looking at each other enables the ICE dynamic clustering algorithm to extract interpersonal gaze. We validate ICE in both video chat using an objective metric with an infrared gaze tracker (F1=0.846, N=8), as well as in face-to-face communication with expert-rated evaluations of eye contact (r= 0.37, N=170). We then use ICE to analyze behavior in two different, yet important affective communication domains: interrogation-based deception detection, and communication skill assessment in speed dating. We find that honest witnesses break interpersonal gaze contact and look down more often than deceptive witnesses when answering questions (p=0.004, d=0.79). In predicting expert communication skill ratings in speed dating videos, we demonstrate that interpersonal gaze alone has more predictive power than facial expressions.
\end{abstract}

}

\maketitle

\IEEEdisplaynontitleabstractindextext

%
\IEEEpeerreviewmaketitle

\IEEEraisesectionheading{\section{Introduction}\label{sec:introduction}}

%
%
%
%
\IEEEPARstart{C}{icero} 
proclaimed "The face is a picture of the mind as the eyes are its interpreter" \cite{mcduff2015crowdsourcing}. Renowned sculptor Hiram Powers stated "The intellect, the will, are seen in the eye" \cite{wright2004henry}. Few us have not heard the adage "the eyes are windows to the soul". Just as ancient history, art, and common wisdom understand the importance of eye gaze, recent scientific research has also begun to demonstrate that eye gaze is a fundamental element of interpersonal communication \cite{argyle1976gaze, chen2002leveraging, macrae2002you}. The importance of eye gaze has also been shown in computer-mediated conversations \cite{fung2015roc, grayson2003you, monk2002look}, as well as in virtual environments \cite{aragon2003creating}. Neuroscientists have even demonstrated that eye gaze behavior is hard-wired into the neural structures of the brain \cite{emery2000eyes}. Despite these findings, many questions regarding eye gaze remain, such as the relevance of eye gaze in detecting deceptive communication.

The advancement in computer vision to automatically extract eye gaze from common video without any special hardware presents new interaction possibilities. It has been estimated that over half of one million smartphone owners use video chat on a regular basis \cite{npdreport} and Facebook alone hosted over 17 billion video chats in 2017 \cite{welch2017}. Further, a recent survey revealed 95\% of large police departments nationwide either have already implemented or are committed to using body-worn cameras \cite{bodycam} to record officers' interactions with members of the public. Video-recording doorbells are also becoming more and more ubiquitous, with market research suggesting that over 22\% of U.S. households will have a smart doorbell by the end of 2019 \cite{smartdoorbell}. The pervasiveness of cameras in our daily lives, paired with the availability of cheap cloud data storage, is creating a previously unimaginable source of video recordings of interpersonal interactions for eye gaze analysis.

As exciting as these developments are, analyzing eye gaze in common video recordings of interpersonal communication remains elusive due to a variety of factors. Current eye gaze detection systems measure eye gaze relative to the video camera (or other dedicated hardware). However, eye gaze relative to a camera tells us nothing about whether the subject is making eye contact with his/her conversation partner. We specifically define an \textit{interpersonal-calibrated} eye gaze signal as a signal which provides a subject's gaze in units of \textit{gaze relative to one's conversation partner}. In order to determine \textit{interpersonal} eye gaze, the physical positions of the camera, the observed subject, and the subject's conversation partner all need to be known. For example, for each of the camera views of individuals engaged in a conversation in Figs. 1a,b,e,f, can you determine whether the participant is looking at his/her conversation partner? 

Even if we extract the eye gaze vectors (shown in red) in each of these images, it does not allow us to answer the question of whether the participants are looking at his/her conversation partner. With contemporary systems, the physical layouts in which the camera recordings were generated, as shown in Figs. 1c,d,g,h, must be known in order to determine interpersonal gaze. Namely, the relative positions of the camera (yellow circles), conversation partners (green circles), and the eyes of the subject whose gaze is being analyzed (blue circles) must be known.

\begin{figure*}[!t]
\centering
\includegraphics[width = \linewidth]{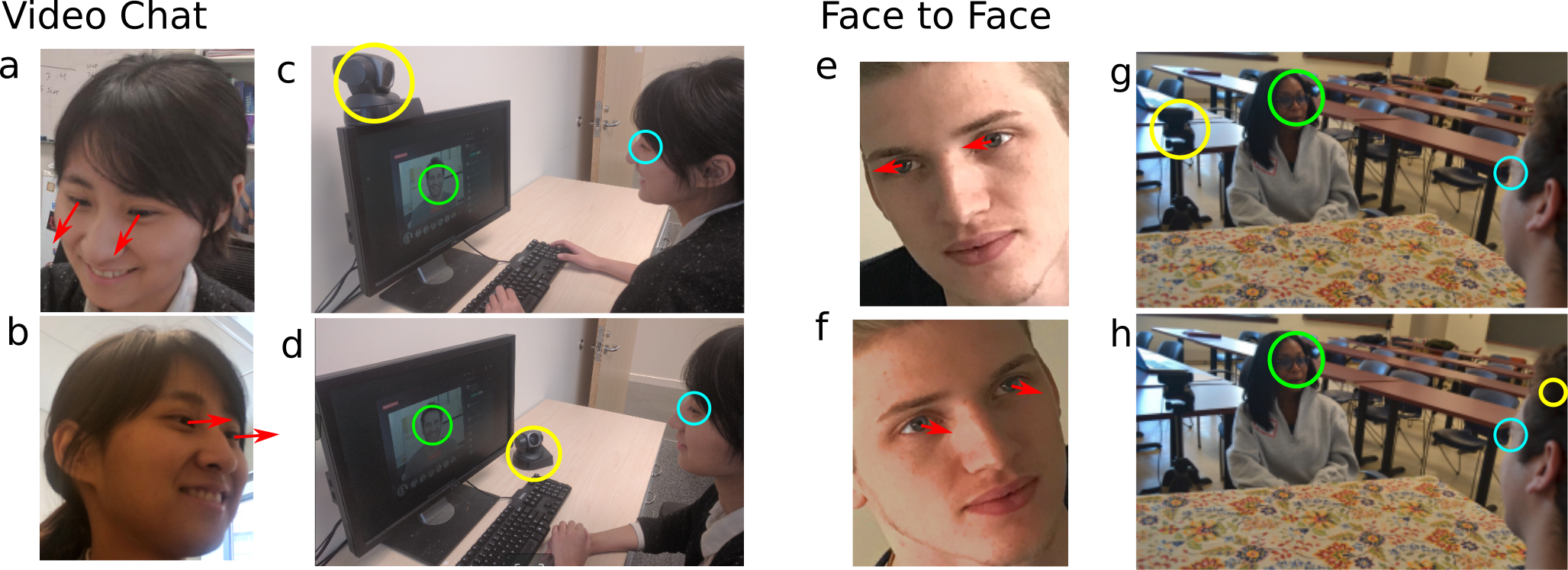}
\caption{\textbf{The Problem of Perceiving Interpersonal Gaze in Unknown Physical Layouts.}  (a,b,e,f: camera view of a participant with gaze vector [red], c,d,g,h: associated physical layouts of the camera [yellow circle], conversation partner [green circle], and subject's eyes [blue circle])}
\label{problemdescribe}
\end{figure*}

While it is seemingly trivial to calibrate a specific hardware and physical configuration for an experiment, it is much more problematic to interpret interpersonal eye gaze from videos recorded from an unknown physical layout. Even when calibration can be performed prior to recording, it is often burdened with the requirement that participants remain stationary or the experimenter continues to re-calibrate. Additionally, when dedicated eye tracking hardware is used, it has been shown to decrease the quality of the findings since subjects tend to make less eye contact with individuals wearing an eye-tracker \cite{Canigueral:2018:DLM:3267305.3274123}. A further difficulty of eye gaze estimations based solely on common video, is that the gaze signals often have substantial levels of noise. 

\begin{figure*}[!t]
  \centering
  \includegraphics[width = \linewidth]{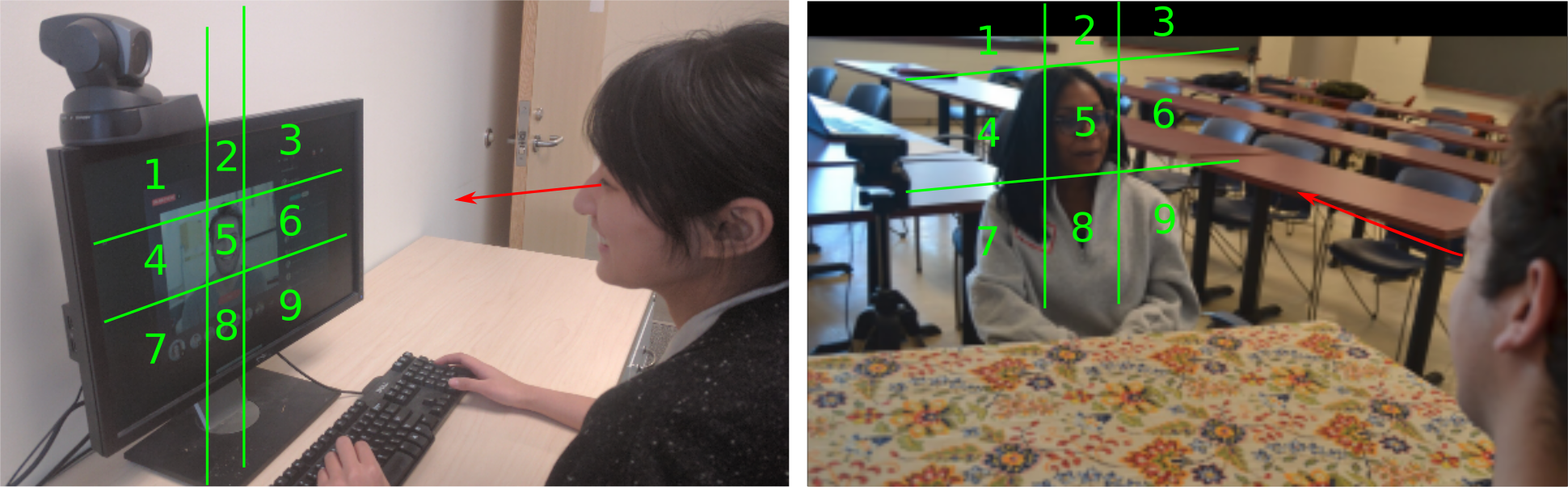}
    \caption{\textbf{Third Person Perspective of the Interpersonal Gaze Regions ICE Predicts.} (Note that this perspective is different than the camera perspective in the recorded videos provided to ICE.)}
  \label{gazeregions}
\end{figure*}

We propose a solution to these problems with the Interpersonal-Calibrating Eye gaze encoder (ICE), which automatically calibrates eye gaze into a discretized, interpersonal gaze-region signal. For example, as shown in Figure \ref{gazeregions}, given a video of a subject from the perspective of the shown cameras, ICE will predict the gaze region (numbered 1 through 9) that the subject is looking at, where the gaze region grid is centered about the subject's conversation partner (i.e. where region 5 is centered over the conversation partner's face). ICE combines automatic gaze extraction with dynamic clustering to automatically convert a camera-relative gaze signal into the a discrete interpersonal signal. 

ICE thus enables the extraction of meaningful interpersonal gaze signal from videos completely agnostic of the position of the recorded subject's conversation partner. This is the case for many of the video sources listed earlier. For example, in common video chat, as shown in Figure \ref{problemdescribe}, the location and orientation of the webcam, the position on the screen where a conversation partner appears, the location of the monitor, the location of the subject are all variables that are typically unknown. Similarly, with police body-mounted cameras, we do not know the distance between the officer's face and the location of the body camera, or even if the officer has rotated his/her torso and thus changed the camera orientation. In the case of smart doorbells, by reviewing the video recording alone, one does not know where the camera has been mounted, or where the conversation partner is located relative to the camera. In all of these cases, with some reasonable limitations, ICE could be applied to extract interpersonal eye gaze.

In summary, we have developed a framework for extracting interpersonal gaze from common video of an individual engaged in dyadic conversation. The remainder of the paper is organized as follows. In the Background section we provide a summary of current state of the art eye gaze extraction systems and their limitations. We then discuss the design motivations in the ICE algorithm and its validation scheme. In the Methods section, we provide the details of the ICE algorithm and the specific approach taken to validate that it is working. The methods section also describes the application of ICE to two separate applications: 1) investigating how eye gaze is relevant to detecting deception communication in a two-player game involving deception over video chat, and 2) determining the relevance of eye gaze in predicting speed dating performance in a face-to-face environment. The Results section describes the validation results as well as the findings of applying ICE to deception detection and speed dating evaluation. The Discussion examines the findings in light of communications and psychological theory, as well as discusses limitations and future work.


Specifically, our contributions include:
\begin{itemize}
\item Development of the novel ICE Interpersonal Calibrating Eye-gaze Encoder framework.
\item Validation of ICE using: i.) an infrared eye tracker with online video chats (F1 score = 0.846 over 88 minutes total involving 8 different participants), and ii) comparison with expert human-rated eye gaze in face-to-face videos (correlation $r$ = 0.37 over 170 rated videos involving 121 participants).
\item Application of ICE to the analysis of eye gaze in face-to-face and computer-mediated communication in two different interpersonal domains: deception detection and communication skills assessment in a speed dating setting. Through these analyses we find that:
\begin{itemize}
\item Honest witnesses in an online interrogation game spend significantly more time looking down relative to his/her conversation partner, compared to deceptive witnesses ($p=0.004$, Cohen's $d=0.794$). This finding supports our intuition that honest speakers tend to break eye contact more often to retrieve a piece of memory in the context of our interaction scenario. 
\item In the automatic assessment of communication skill, we find that interpersonal gaze has more predictive value than averaged facial action units (i.e. automatically coded facial expressions).
\end{itemize}
\end{itemize}

 

\section{Background}
In this section we provide an overview of gaze extraction technologies and describe their limitations in solving the problem of determining interpersonal gaze in an unknown environment (Figure 1). We then demonstrate the value of interpersonal eye gaze by describing multiple domains in which interpersonal gaze has been shown to be important which we will apply ICE to. The section ends with the design motivations in developing and robustly validating the ICE algorithm.

\subsection{Gaze estimation methods}
Kar et al. provide a survey of remote gaze estimation techniques \cite{kar2017review}. These techniques fall into two main classes depending on whether they are based on infrared light or visible light. Infrared-based techniques are commonly referred to as PCCR (pupil center and cornea reflection) methods. Techniques which use visible light rely on eye-local visual features and/or shape to estimate gaze direction. These are commonly referred to as appearance and shape-based methods. 

\subsubsection{PCCR-based gaze estimation}
Often referred to as the gold standard of eye tracking due to their accuracy, PCCR-based methods require specialized hardware for both illuminating a subject's eyes with infrared light, and recording its reflection\cite{guestrin2006general}. PCCR-based methods determine a subject's eyeball orientation relative to the infrared camera through use of a variety of mathematical models of how the infrared light should reflect. These models include 2D regression \cite{blignaut2013mapping, cherif2002adaptive}, the 3D model \cite{meyer2006single}, and other techniques, each with their differing advantages and weaknesses. The 2D regression method maps the vector between pupil center and corneal glint to coordinates on the monitor using a polynomial transformation function \cite{blignaut2013mapping, cherif2002adaptive}. The 2D regression method, while computationally efficient, requires calibration, and is vulnerable to head movements. While this method has been made more robust with neural networks \cite{jian2009eye, wang20162d}, their accuracy still depends on people keeping their heads in the same position throughout the whole gaze estimation procedure. The 3D model technique tries to achieve better accuracy by modelling the human eye geometrically to estimate the center of the cornea along with the optical and visual axes of the eye \cite{meyer2006single, hennessey2006single, guestrin2006general}. While each of the PCCR-based eye tracking systems introduced above require human-involved calibration for accurate gaze tracking, the recent PUPIL system removes the need for user-involved calibration with the introduction of a specialized near-infrared headset \cite{kassner2014pupil}. 


Despite the existence of highly accurate PCCR-based gaze tracking, some of which even allow automatic calibration, none of the described systems directly measure interpersonal eye gaze. A prior knowledge of where an individual's conversation partner is in real life (for person to person recordings) or on the screen (for video chat conversations) has been necessary if one is to directly convert provided eye gaze into interpersonal gaze.

In summary, PCCR-based infrared eye gaze trackers 
\begin{itemize}
\item{require specialized hardware}
\item{often require calibration}
\item{suffer in accuracy if head pose is not maintained}
\item{provide eye gaze in absolute coordinates, not interpersonal gaze coordinates}
\item{require knowledge of an individual's conversation partner location if eye gaze will be converted directly to interpersonal gaze coordinates}
\end{itemize}

\subsubsection{Appearance and shape-based gaze estimation}
Appearance and shape-based gaze estimation try to obviate the need for specialized hardware by extracting gaze from common video. 
Shape-based techniques attempt to extract gaze through the use of deformable templates of the eye area to which an eye image is fit \cite{reinders1997eye, ramadan2002eye, wang2011driver, ince20112d}. The deformable templates involve a generic representation of an eye, such as two parabolas describing the eye contours and a circle for the iris itself \cite{kar2017review}. Then, depending on the image of the eye, the parameters can be adjusted to fit a person's unique eye appropriately. The shape-based method is prone to have accuracy suffer with head pose variations 
since different head poses sometimes occlude parts of the eyes and reduce the tracking accuracy of shape-based methods.
In addition, shape-based methods may run into instances in which a participant's eye image does not fit well into one of the model's templates, as well as computational complexity issues that arise in needing to adapt to extremely variable eye shapes.

Appearance based techniques work by extracting information from the eye region using representative models trained on features from a vast quantity of eye images \cite{baluja1994non, bacivarov2008statistical}. Modern appearance based methods employing SVM classifiers, Local Binary Pattern (LBP), LBPH and PCA in order to improve accuracy \cite{wu2014gaze, lu2008gaze, yilmaz2016local}, with state of the art systems utilizing CNN-based deep learning approaches \cite{george2016real}, \cite{zhang2015appearance}. Appearance-based methods have attempted to further reduce gaze errors due to head pose by combining their output with head pose detection algorithms \cite{wood2015rendering}. However, even when tuning models with large data sets, the overall accuracy has generally been lower than PCCR-based methods.

Although appearance and shape-based techniques are less accurate compared to PCCR methods in general, they do not require specialized hardware (beyond a video camera), nor do they require calibration for detecting gaze relative to the camera. However, like PCCR methods, appearance and shape-based techniques also do not directly determine interpersonal gaze, i.e. these methods also need knowledge of the physical location of the observed subject's conversation partner in real life or on video chat screen.

In summary, Appearance and shape-based methods 
\begin{itemize}
\item{only require basic video camera hardware}
\item{do not require calibration (but only provide eye gaze in camera coordinates vs real world coordinates without calibration)}
\item{suffer in accuracy unless head pose variable is not explicitly accounted for}
\item{provide eye gaze in camera coordinates, not interpersonal gaze coordinates}
\item{require knowledge of an individual's conversation partner and the camera position if eye gaze will be converted directly to interpersonal gaze coordinates}
\end{itemize}

\subsubsection{Other gaze estimation techniques}
Beyond eye gaze extraction methods involving infrared eye trackers and camera-based gaze extraction, other gaze extraction technologies involving other specialized hardware have been developed. Ye, et. al, have attempted to detect eye contact using specialized glasses which contain both front-facing and self-facing cameras to record both the wearer's gaze direction, as well as video of what the wearer sees. With their hardware, Ye, et. al report achieving a frame-based eye contact detection precision of 80\% and recall of 72\%\cite{ye2012detecting}. Another recent method developed by Bulling, et al., quantifies eye gaze using electrooculography to measure the brain waves associated with specific types of eye-gaze activity \cite{bulling2011eye}. Bulling and his team developed algorithms to detect three specific eye-movements (saccades, fixations, blinks) based off brain activity. While this technique does not provide eye gaze directly, the utility of the their encoding scheme was shown in classifying five different activities (copying a text, reading a printed paper, taking handwritten notes, watching a video and browsing the web) \cite{bulling2011eye}. While this technique interprets neurological signals, it also does not directly provide an interpersonal gaze signal.

In summary, to the best of our knowledge, existing eye gaze tracking technologies do not provide a direct way to capture interpersonal eye gaze from common video without knowledge of the associated physical layout of the camera position and the location of a conversation partner.



\subsection{Interpersonal Eye Gaze}

Prior studies have investigated interpersonal eye gaze behavior in different domains, including common video chat, deception detection, and communication skills assessment. Recognizing the importance of eye gaze and the detrimental effect of when it is broken over common video chat, research has addressed the problem of the eye contact "parallax". eye contact parallax refers to the condition in which a video chat participant looks at their counterpart directly on his/her computer screen, but since the participant's video camera is arranged above their computer screen, it appears as though eye contact is not being made. Solina \& Ravnik  attempt to correct perception of this parallax using a Mona Lisa effect-type of perceptual trick with an adjusted perspective and a tilted monitor \cite{solina2011fixing}. Others have developed techniques of directly morphing live video images in order to correct eye contact parallax using an predetermined camera perspective \cite{kuster2012gaze}.

Building off the findings that eye gaze is an indicator of human behavior, interpersonal eye-gaze has been studied in depth in regards to eye contact in deceptive communication with contradictory results. Studies have shown that people are gaze-avoidant when they feel ashamed \cite{ekman2009telling, mehrabian2017nonverbal}, and that people feel ashamed when they engage in lying \cite{depaulo2003cues}. Past research has also demonstrated that people will frequently break gaze when their cognitive load increases \cite{doherty2002development, doherty2005gaze} and that lying is associated with higher level cognitive processes \cite{vrij2006detecting, vrij2008cognitive, vrij2011outsmarting}. Vrij et al. used recalling an event in reverse chronological order to illustrate this point \cite{vrij2008increasing}. While these findings support the notion of gaze avoidance during deception, other research has come to different conclusions. For instance, it was demonstrated that people attempt to persuade others by looking them in the eyes \cite{kleinke1986gaze}, with findings suggesting that eye contact makes individuals appear more trustworthy \cite{bayliss2006predictive}. Studies directly measuring eye contact in deceptive communication have both found eye contact to be increased \cite{sitton1981detection, bond1985miscommunication} as well decreased \cite{granhag2002repeated} among liars. Given the high dimensional nature of lying (i.e. unique situations, unique individuals, unique stakes), it is entirely likely that depending on the context, results on gaze may differ, which remains an unanswered question. 

\subsubsection{Communication skills, eye gaze meaning, assessment and training}
The concept that eye gaze behavior may indicate different things in different contexts has also been suggested in communication skills assessments. Communication skills can be broadly defined as a measure of how competent a person is in social situations. 


Schilbach demonstrated the crucial importance of gaze in social interactions and through analyzing the dynamics of gaze in different contexts, proved that the context changes how gaze should be interpreted \cite{schilbach2015eye}. The fact that gaze interpretation is context-dependent indicates that gaze must be studied in a variety of social situations for both in person and virtual environments to fully understand its meaning. Past work has utilized virtual environments to research eye gaze \cite{bailenson2001equilibrium} and eye gaze has been used in virtual environments to gauge user attention and facilitate more natural human-to-computer interactions \cite{peters2010investigating}. However, use eye gaze as a feature to help evaluate communication skills in automated conversational coaching systems remains an unsolved problem. Although there are systems available to coach a person in communication skills, such as public speaking, that use non-verbal behavior measurements to provide feedback on improving \cite{fung2015roc}, interpersonal gaze analysis cannot be offered as a feature without specialized hardware and environmental settings given the previously stated problems that give rise to gaze estimation inaccuracies. Systems have also been developed that employ a virtual agent to interact with a person to build his/her generic conversational skills \cite{ali2015lissa}, as well as conversational skills specific to groups of people (i.e. older adults) \cite{ali2018aging} and for certain contexts (i.e. job interviews) \cite{hoque2013mach}. However, none of these systems analyze gaze patterns. 

\subsection{Unsupervised Gaze Target Classifiers}
Prior work has been done on the development of unsupervised classification of when an individual is looking at a "target" using only standard video recordings of in-person interactions. Zhang et al. presented an unsupervised algorithm for detecting when an individual looks at a target located close to the video camera \cite{zhang2017everyday}. The algorithm assumes that an individual's gaze will form a cluster close to the origin (i.e. the coordinates of the camera). In order to identify this cluster, they apply the OPTICS clustering algorithm \cite{ankerst1999optics} to raw eye gaze signals extracted from the standard video.  Frames which are in the cluster closest to the origin are labelled as being "on target" and are used to train an SVM classifier which predicts "on target" gaze from facial landmark features. Zhang et al. suggest using three hours of data to train the SVM. Their method is validated on human-annotated data to yield a Matthews Correlation Coefficient (MCC) of between 0.2 (for 30 minutes of training data) to 0.6 (for four hours of training data).

Attempting to overcome Zhang et al.s requirement that the camera is placed next to the target, Muller et al. introduced a non-clustering algorithm for detecting "on person" gaze in a multi-person conversation setting \cite{muller2018robust}. Their algorithm leverages the observation that individuals tend to look at the person who is talking. In this regard, Muller et al. set up eight video cameras, two behind each of four speakers arranged in a conference type of setting, to use for automatic facial expression extraction to determine who is speaking in each frame \cite{muller2018robust}. The speaker labels are then used to train a multiclass SVM to predict who a person is looking at among the other speakers present. Portions of data in which a given participant is speaking are thus not used to train his/her gaze classifiers (i.e. only "listening data" is used). Validation was done using human-judged ground-truth labels in which the annotators determine which of the three other participants a given speaker is looking at (or whether they are not looking at any of the others) from the video recordings. Muller et al. report a calibration time of 6 minutes, with an accuracy of 70\%.


Given the inability and/or unknown performance of current systems to measure interpersonal gaze from video chat, coupled with both the need to advance our understanding of interpersonal gaze in differing contexts, there exists a strong need for a system which can automatically extract interpersonal gaze from common videos. In light of the prior work, we hope to design an eye gaze encoder for standard video recordings which: 1) works on both video chat as well as in-person recordings, 2) is applicable to short recordings (i.e. does not require substantial calibration time), 3) provides a directional measure of interpersonal gaze (as opposed to Boolean target detection) and 4) does not require any assumption about the physical layout.

\subsection{Background Design Considerations in Designing and Validating ICE}

A fundamental principle of an individual's eye gaze is that gaze is concentrated in a number of regions or clusters (i.e. looking at particular areas), and that for individuals in a conversation, the most commonly looked at region is one's conversation partner \cite{vertegaal2001eye}. More specifically, using infrared eye tracking, Vertegaal, et al., found in their study of face-to-face conversations, participants looked at their conversation partner with a probability of 77\% when they are listening, and 88\% when they are talking \cite{vertegaal2001eye}. A motivating principle in the design of the ICE algorithm is to determine where a recorded subject's conversation partner is by detecting dense regions of eye gaze. While clustering algorithms are well suited to identifying dense regions, the clustering algorithm employed should be able to gracefully handle outliers, and should be able to handle non-specific shaped regions (since an individual may tend to look either at the others eyes only, whole face, or the conversation partner may even move to some extent). The clustering algorithm should also automatically support a wide range of contexts i.e. different physical settings (video chat and face-to-face) as well as different recording distances. These different contexts may need differing cluster parameters, preferably supported by an automatic parameter selection method. In validating the designed system, it is important that the validation technique involves both face-to-face and video chat based methodologies, since it is likely that good performance in one domain might not indicate good performance on another.

\section{Methods}
\subsection{The ICE algorithm}

\begin{figure}[h]
  \centering
  \includegraphics[width=0.9\linewidth]{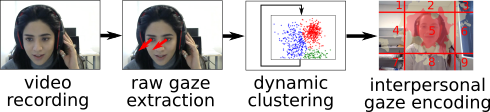}
  \caption{ICE Framework for encoding interpersonal eye gaze from raw video.}
  \label{flowchart}
\end{figure}

In this section, we describe the ICE framework and go into detail of our proposed method for extracting interpersonal eye gaze from common video. Shown in Figure \ref{flowchart} are the high level steps of the ICE encoder framework. In summary, the framework involves starting with a video recording, which is then analyzed by an eye gaze extraction tool. The extracted raw gaze signals (x and y), are then provided to the dynamic clustering algorithm, which outputs interpersonal-calibrated encoded (ICE) gaze signals. For each video frame, a single gaze region number (1-9), is produced.

\subsubsection{Automatic Gaze Extraction}
In order to extract the raw eye gaze values from recorded videos, we used the OpenFace open source tool \cite{baltrusaitis2018openface}. OpenFace was selected due to its open source nature, public benchmarks, and reasonable performance. However, any reliable tool for extracting eye gaze from video can be used with our algorithm. It is important to note that OpenFace eye gaze values are in world coordinates (i.e. the x and y gaze angles provided are in radians relative to the camera position). It should be emphasized that these raw gaze angles are not in units relative to the position of the participant's conversation partner.

\subsubsection{Dynamic Clustering}
We use the DBSCAN density-based clustering \cite{ester1996density} on the raw eye gaze angles to identify dense regions of eye gaze. DBSCAN is especially appropriate, since unlike most clustering algorithms, it allows points to \textit{not} be in a cluster by default. For example, algorithms such as k-means or Gaussian Mixture Model will have a hard time reconciling the data if a large number of gaze positions are seemingly sporadically distributed. Additionally, DBSCAN is used due to its ability to detect gaze clusters which may not be perfectly circular in shape. 
Most importantly, DBSCAN is particularly suitable to be applied to eye-gaze data since eye gaze signals typically form dense regions representing objects that users look at during recordings.
Most, if not all, clustering algorithms are heavily influenced by their parameter selections. This is also true of DBSCAN, which is sensitive to its two input parameters $\epsilon$ and minPts\cite{ester1996density}. $\epsilon$ defines the maximum distance between two points for them to be considered in the same neighborhood and minPts defines the minimum number of samples in a neighborhood required for a point to be considered a core point.

\begin{itemize}
    \item minPts estimation: The minPts parameter loosely correlates with the resulting number of points in a cluster. Therefore, using a small value of minPts would return clusters with only a few data points in it. We utilize this property to avoid instability due to noisy eye gaze extraction. Specifically, since OpenFace extracts eye gaze at the frame rate (which is 15 to 30fps in typical video), and since the process is noisy, we need to ensure that small clusters are not formed around noisy regions. To help avoid this scenario, we set the minPts to be 1 percent of the total number of extracted frames for each video so that the output clusters are of reasonable size in comparison to the data.
    \item $\epsilon$ estimation: Due to the freedom the DBSCAN algorithm has in defining cluster regions with non circular shapes, it is crucial to have a good $\epsilon$ value to model interpersonal gaze appropriately. For example a small $\epsilon$ value could result in a cluster being defined inside another cluster, while setting a large $\epsilon$ value could inappropriately merge dense regions together, resulting in the loss of useful information. One of the most commonly used methods to select the parameter in DBSCAN is to use a k-distance plot to determine $\epsilon$. However, this method typically requires human effort in manually selecting $\epsilon$ at the "knee" of the plot \cite{sander1998density, kassambara2017practical, sawant2014adaptive}. To automate the process of choosing $\epsilon$ we leverage our domain knowledge that there is one person appearing on the screen and taking up most of the screen, which would account for a majority of the eye gaze frames in the OpenFace data. In other words, we want to detect two clusters, one main cluster and one outlier cluster. Therefore, we perform a sequential grid search for the appropriate $\epsilon$ value, taking the maximum value for which a single dominant cluster is formed (i.e. $\epsilon$ is initialized to be 1). Specifically, the value of $\epsilon$ starts from a relatively large value for the data, causing DBSCAN to return one output cluster. Then, the value of $\epsilon$ is reduced gradually until the desired number of clusters (two in this case) appears.  In defining the condition of when there is a single dominant cluster, we require that the ratio between the two largest clusters (which includes the outlier cluster) is less than 10. The ratio is set to 10 based on empirical evidence in that: if this ratio is set to a large number, it is possible that some data points around the boundaries of the clusters are not labelled correctly due to the small size of the newly formed cluster(s); and if this ratio is set to a small number, the algorithm might not be able to find the two clusters and return "FAIL".
\end{itemize}

\subsubsection{Re-normalization and Encoding}
After running DBSCAN on the data, we define the rectangular bounding box of the largest cluster as the Region of primary Visual Engagement (RVE). The RVE should represent the region of an individual's conversation partner as long as the fundamental principle that the densest gaze area occurs when looking at the person an individual is conversing with.
\begin{figure}[htp]
\fbox{
\begin{minipage}[b]{0.9\linewidth}
\textbf{DEFINITIONS} \\ 
$\textit{frames}$ = raw gaze angles (x and y) at 15fps \\
$\epsilon$, $minPts$ = parameters for DBSCAN clustering\\
DBSCAN = apply the DBSCAN algorithm \\
$A[i]$ = sequence of cluster assignments from DBSCAN for frame i \\
$C[k]$ = number of frames assigned to cluster k in A \\
$k'$ = largest cluster in A (i.e most likely defined by the facial region of the conversation partner) \\
$X$ = bounding box of $k'$ \\
---------------------------------------------------------------------\\
\begin{algorithmic}[1]
\STATE $minPts$ = length(frames) / 100 \\
\STATE initialize $best\epsilon$ = 0\\
\STATE initialize $\epsilon$ = 1 \\
\WHILE{$\epsilon >= 0$}
\STATE A = DBSCAN(frames,$\epsilon$, minPts) \\
\STATE B = set of unique elements in A \\
\IF{$length(B) < 2$}
    \STATE $\epsilon$ = $\epsilon$ - 0.01 
    \STATE \textbf{continue}
\ENDIF
\STATE $Csort$ = sort C in descending order
\IF{$\frac{Csort[0]}{Csort[1]} <= 10$ \textbf{and} largest cluster is not the outlier cluster}
    \STATE $best\epsilon$ = $\epsilon$ 
    \STATE \textbf{break}
\ELSE 
    \STATE $\epsilon$ = $\epsilon$ - 0.01
    \STATE \textbf{continue}
\ENDIF
\ENDWHILE
\IF{$\epsilon < 0$}
    \STATE $best\epsilon = 0.001$ 
\ENDIF
\STATE A = DBSCAN(frames, $best\epsilon$, minPts) \\
\STATE B = set of unique elements in A \\
\STATE find $k'$ then find $X$ \\
\STATE calculate $output\_sequence$ \\
\IF{$length(B) > 1$}
    \RETURN{$output\_sequence$}
\ELSE 
    \RETURN{FAIL}
    \COMMENT{ cannot find more than 1 cluster}
\ENDIF
\caption{Dynamic Density-based Clustering}
\end{algorithmic}
\end{minipage}
}
\end{figure}
\begin{figure}[!hbp]
  \centering
  \includegraphics{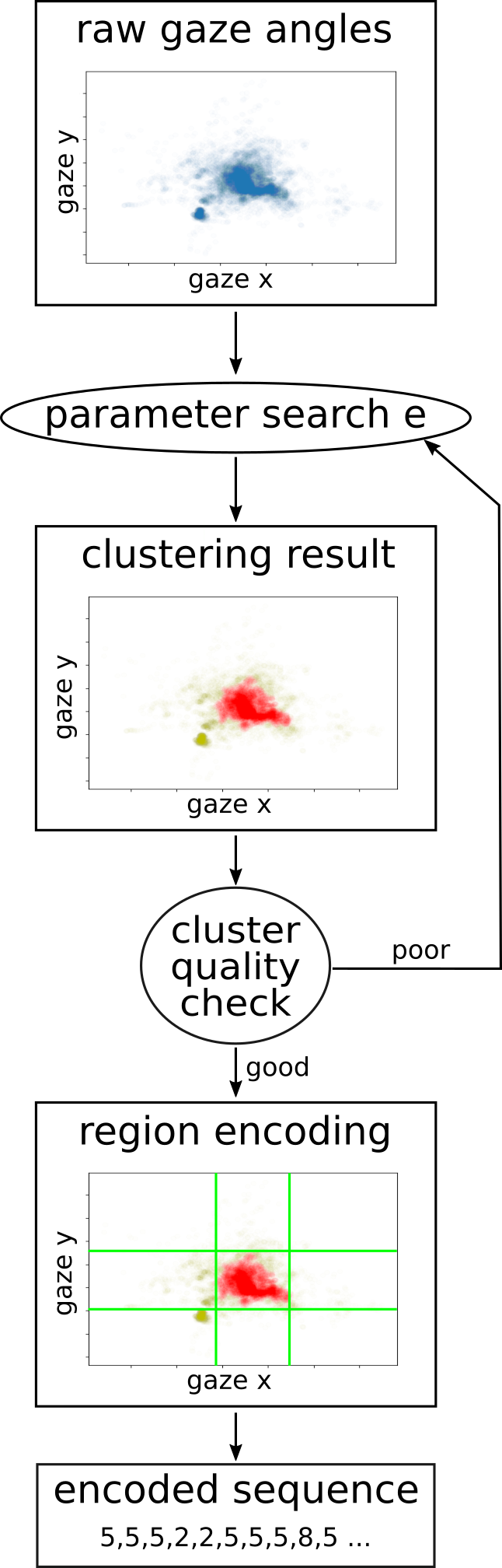}
  \caption{Block diagram of the encoding procedure}
  \label{triple}
\end{figure}
After defining the RVE region for the data, we encode all the data points into one of the nine regions of a 3x3 grid with the RVE as the center region (i.e. region \# 5 when numbering the regions 1 through 5 starting at the top and going to the right). Specifically, the encoder is constructed based on a bounding box containing the RVE region, with the boundaries defined by the minimum and maximum raw eye gaze values in each dimension of the data. Then, by extending the boundaries of the bounding box 9 discrete regions are created, corresponding to the nine possible direction of eye gaze with the center region representing the RVE region (i.e. making gaze at conversation partner).

\subsection{Validation}
\subsubsection{Infrared Eye Gaze Tracker Validation in Video Chat}
We test the accuracy of the eye calibration algorithm with an infrared Eye Gaze tracker. Specifically, we record videos of conversations over video chat of two individuals playing an interrogation-based game with the addition of an infrared eye tracker (GP3) \cite{gazepoint} to accurately track the eye gaze of participants. For the signals output from the infrared eye tracker, we select the bounding box containing exactly the face of the participants' conversation partners as the ground truth of the RVE region for the comparison. Because the signals from OpenFace and the infrared eye tracker are independent, it is important to synchronize the two signals. In this regard, we perform a convolution between the signal of the up-down dimension in GP3 with the y-signals (up-down direction) of OpenFace. The up-down direction is chosen over the left-right direction because it has more variation, allowing the convolution to better detect the point of synchronization. The raw OpenFace signal (15fps) and GP3 signals (60fps) were downsampled to 3 fps. We then convert the gazepoint signal into 9-region encodings, by defining a 3x3 grid, with the center rectangle being the bounding box of the largest cluster (RVE).

The validation process for one of the video chat interactions with the infrared eye tracker is shown in Figure \ref{triple}. Figure \ref{triple}a) shows the resulting convolution between the downsampled Gazepoint x-output and the downsampled OpenFace gaze x-angle. Note that the maximum amplitude occurs at a phase shift between zero and 300 seconds. Shown in \ref{triple}b) are the Gazepoint and OpenFace x-signals shifted by the phase shift with the maximum amplitude in Figure \ref{triple}a). Note in Figure \ref{triple}b), the large increase in both signals at approximately 60 seconds corresponds to the time in which the display image disappears and the video windows activate, thus, showing the gaze shifting. Figure \ref{triple}c) represents the heat map of the gaze point activity overlayed with the resulting ICE encoded regions. 

\begin{figure*}[hbt]
  \centering
  \includegraphics[width=0.9\linewidth]{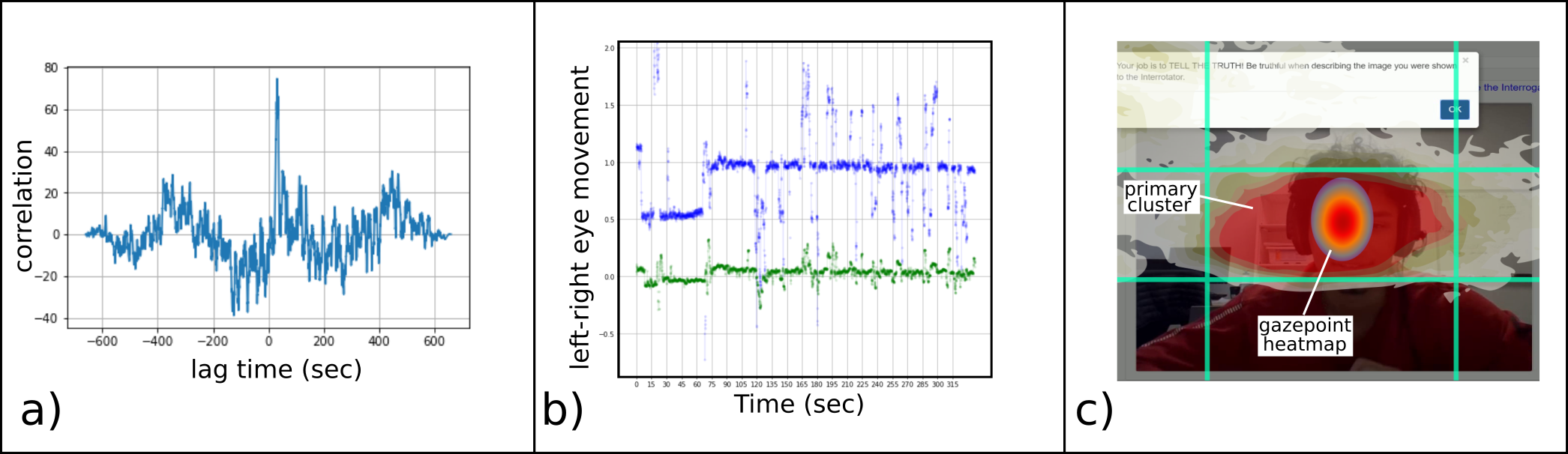}
  \caption{\textbf{Validation of ICE gaze region output with GP3 Infrared-Eye tracker} (a) graph of correlation of OpenFace and GP3 x-axis signal for determining optimal phase shift for signal synchronization; (b) scatter plot of OpenFace gaze x-axis (green) and GP3 gaze x-axis (blue) after synchronization; (c) comparison of GP3 heat-map and ICE encoder cluster result.}
  \label{triple}
\end{figure*}

\subsubsection{Expert-Rated Eye Contact Validation in Face-to-Face Videos}
To validate the performance of the encoder on a face-to-face situation, with the camera placed in a third-person perspective as shown in Figure \ref{problemdescribe}, we apply the encoder to a speed dating dataset \cite{ali2015lissa}. Specifically, for the recording of each participant, we obtain the encoder outputs and calculate the proportion the individual looking into the RVE. 

In the speed dating study, trained Research Assistants (RA) gave ratings on some criteria to participants after a short face-to-face conversation of 4 minutes. One of the ratings is conversational skills. The rating scheme was taken from the well established conversational skills rating scale (CSRS) \cite{spitzberg1988communication}. The rating is based on appropriate use of nonverbal cues, which includes effective eye contact. Therefore, it is reasonable to assume that a measure of eye gaze will be able to explain the variance of the skills ratings to some extent.

\section{Cases of Study}
To show the usefulness of the algorithm, we applied the encoder to two datasets of dyadic videos: the deception dataset and the speed dating dataset. In the deception dataset, we assess ICE's ability to analyze gaze patterns in the context of deception using a computer mediated platform. In the speed dating study, we use ICE to analyze the role of gaze patterns in a face-to-face speed dating study. We wanted to explore if ICE can work across multiple interaction scenarios and platforms. A summary of the two datasets is given in Table \ref{table_dataset_summary}.

\begin{table}[]
    \centering
    \caption{Summary of datasets.}
    \resizebox{\columnwidth}{!}{
    \begin{tabular}{|c|c|c|} 
        \hline
        Dataset & Deception & Speed Dating\\
        \hline
        \# samples & \makecell{95\\(8 for validation)} & 170 \\
        \hline
        \makecell{duration \\ average (min) \\ total (hours)} & \makecell{\\ 10\\ 15.8\\} & \makecell{\\ 4\\ 11.3\\} \\
        \hline
        protocol&ADDR\cite{sen2018automated}&Wizard of Oz\cite{ali2015lissa}\\
        \hline
        ground truth & IR camera & human rating\\
        \hline
        camera placement & \makecell{various webcam positions,\\different recording environments}& \makecell{behind and to side\\ of RA(s)}\\
        \hline
    \end{tabular}
    }
    \label{table_dataset_summary}
\end{table}

\subsection{Deception Detection}
\subsubsection{Study Data}
The dataset that we used for deception comprises 87 dyads of voluntary deception. One data sample is a pair of video recordings where the individuals participate in a communication experiment involving deceptive behavior. All data samples are collected using the ADDR framework \cite{sen2018automated}. In the experiment, one participant is the interrogator and the other is the witness (roles randomly assigned). The experiment consists of two phases. In the first phase, the interrogator asks the witness baseline questions and in the second phase, the witness is shown an image (the evidence) for 60 seconds and instructed to memorize that image. The witness then makes a decision to tell the truth about the image they previously saw, or decides to bluff about seeing a different image. The evidence a witness could be shown comes from a database comprised of image and word description pairs. To help ensure proper participation in the experiment, the experimenters implemented an incentive system of bonuses. A bonus is awarded to the interrogator if they are able to correctly identify the honest or deceptive behavior of the witness, while a bonus is awarded to the witness if the interrogator believes them. To raise the stakes of deception, the protocol instituted a random bonus round worth 50 dollars decided by the roll of a die (causes participants to treat every round as if it could be worth 50 dollars). For that high-stakes round, if the interrogator is correct as to whether the witness was bluffing or telling the truth they are awarded 50 dollars. If the witness convinces the interrogator that they were telling the truth for the high-stakes round, they are awarded 50 dollars. 
\subsubsection{Analysis with ICE Encoder}
We applied the encoder to the deception dataset containing 87 dyadic videos of individuals playing a game as described (47 recordings contain voluntary bluffers and 38 recordings contain voluntary truth tellers). The nine relative frequencies of the nine regions outputted from ICE are calculated for each video. These frequencies are then used for statistical analysis and classification of deceptive behaviors.
\begin{itemize}
    \item Statistical analysis: In order to compare the frequencies of the gaze regions between truth tellers and bluffers, we use the independent t-test. Cohen's d is used to measure the differences between the distributions in estimated standard deviations (i.e. effect size)\cite{cohen1988statistical}. 
    \item Classification: We next consider the usefulness of the introduced ICE gaze statistics by measuring whether classification of deceptive and honest communication can be improved by using average ICE region statistics. More specifically, we use logistic regression to classify deceptive and honest communication with three different feature sets:  ICE gaze region frequency, affective facial expression features, and the combined ICE frequency and affective facial features. For affective facial expression feature extraction, we use emotion features provided by the Affdex facial analysis tool (including: joy, fear, disgust, sadness, anger, surprise, contempt, valence, and engagement) \cite{mcduff2016affdex}. Both l1 and l2 regularization are used, with the $\lambda$ hyperparamter determined using a dev set and selecting the parameter which gives the best cross-entropy. Since study participants may have played the game multiple times, we also ensure that same individual does not appear in the training, validation and test set when splitting the folds to better assess the quality of the result.
\end{itemize}

\subsection{Communication Skill Assessment in Speed Dating}
\subsubsection{Study Data}
To test our frameworks efficacy we applied ICE to another dataset. This dataset was collected from a speed dating study \cite{ali2015lissa, ali2018and}. The aim of the speed dating study was to validate an online virtual assistant based social skills intervention in a randomized control setting. The dataset contains face-to-face interactions in a speed dating session (i.e. casual getting to know each other conversation) between the study participants (n=23) and female research assistants (n=8). Each participant was male and interacted with 3-5 trained female research assistants in two separate sessions (pre and post intervention). Each conversation was exactly four minutes long. We collected 170 videos of conversation in total. Each participant's interaction was rated by their partner using the well established conversation skills rating scale (CSRS) \cite{spitzberg1988communication}. The ratings include level of eye contact and overall communication skill. Unlike the deception dataset, each video was recorded from a third person perspective shown in Figure \ref{problemdescribe}gh.

\subsubsection{Analysis with ICE Encoder}
We first extracted the action units using Openface \cite{baltrusaitis2018openface} and eye contact features using the ICE framework. We applied linear regression with L1 regularization (LASSO) on the extracted features to predict the conversational skill ratings. Specifically, to demonstrate the value of ICE output, LASSO is used with the inputs being the average of Action Units along with the proportion of times a person spends looking at his conversation partner. The regularization hyper-parameter of LASSO is picked by optimizing the root mean square error.

As another evaluation of the predictive power of ICE to participants ratings, we compare the mean squared errors of LASSO with two different inputs. One input containing just the average of Action Units and the other input containing the average of Action Units with the additional feature of the proportion of time a person spend looking at his conversation partner (frequency of the RVE region). The inputs are all regularized to ensure the features are all on the same scale. We also perform a 5-fold cross validation with a dev set to choose the hyper-parameter for the regression model. 


\section{Results}
\subsection{Validation}
 
\subsubsection{Infrared Eye Gaze Tracker Validation in Video Chat}
Of the ten videos for which ICE was compared with the infrared eye gaze tracker, two of the videos could not be used since the participants moved too far out of the eye tracker's calibration zone (and the video camera region) during the video chat recording. Shown in Table \ref{infrared_validation_accuracy}, is the ICE accuracy for each of the validation videos in correctly predicting on-person interpersonal gaze (RVE). These numbers represent the percentage of downsampled frames (3 fps) that ICE matched the on-person eye gaze determination from the infrared eye gaze ground truth. The average accuracy and F1 scores over the eight videos is \textbf{76.6\%} and \textbf{0.846} respectively, with corresponding standard deviations of 11.8\% and 0.0857. 

\begin{table}[]
    \centering
    \caption{Accuracy of ICE interpersonal gaze in comparison to infrared eye tracker ground truth.}
    \begin{tabular}{|c|c|c|} 
        \hline
        file \# & ICE-RVE accuracy & F1-score\\
        \hline
        1 & 0.916 & 0.953 \\
        \hline
        2 & 0.702 & 0.816\\
        \hline
        3 & 0.951 & 0.974\\
        \hline
        4 & 0.774 & 0.863\\
        \hline
        5 & 0.740 & 0.829\\
        \hline
        6 & 0.661 & 0.744\\
        \hline
        7 & 0.611 & 0.738\\
        \hline
        8 & 0.743 & 0.849\\
        \hline
        \hline
        Average & 0.762 & 0.846\\
        \hline
        St. Dev. & 0.118 & 0.0857 \\
        \hline
    \end{tabular}
    \label{infrared_validation_accuracy}
\end{table}


\subsubsection{Expert-Rated Eye Contact Validation in Face-to-Face Videos}
The average \% of frames in RVE provides ICE's estimation of the proportion of time a person (participant) looking at the conversation partner (Research Assistant). On the other hand, the research assistants are asked to evaluate participants eye contact scores, so their ratings should be a reflection of how often the participants look at them. Although the RVE is not the region of the eyes only, a study showed that people can hardly distinguish whether a person is looking into their eyes versus other parts of their faces \cite{lord1974perception} Therefore, if ICE estimates correctly, it is expected that the average percentage of frames in RVE correlates with human ratings.

Of the 170 face-to-face video recordings with expert human-rated eye contact labels, ICE failed to properly detect eye gaze on 4 videos. For each of these videos, the algorithm detected only a single cluster with all the data points (i.e. the algorithm was not able to distinguish a separable dense region of eye gaze).
A comparison of ICE with human expert eye contact ratings (ranging from 1 to 6) is provided for the remaining 166 samples (out of 170) in Table \ref{speedating}. More specifically, in creating this table, first the \% of frames for which ICE determines the individual is looking at their conversation partner (the region of visual engagement - RVE) were calculated for each of the 170 videos. Then, for each row of the table, the average RVE \% is reported as an average over all videos which received a specific rating. As shown in Table \ref{speedating}, the average RVE \% increases with the human-expert labeled eye contact rating. The correlation between the human labeled eye contact ratings and the averaged RVE is $r = 0.917$, whereas the correlation between the human labels and RVE on a per-recording basis is $r = 0.37$. 

\begin{table}[]
    \centering
    \caption{Comparison between human labeled eye contact ratings and mean ICE gaze encoding output}
    \begin{tabular}{|c|c|c|} 
        \hline
        \makecell{eye contact \\ rating} & \makecell{\# rated \\ videos} & \makecell{average \% of \\frames in RVE} \\ [0.5ex] 
        \hline
        1 & 2 & 76.4\\
        \hline
        2 & 7  & 82.0\\
        \hline
        3 & 38 & 88.6 \\
        \hline
        4 & 45  & 87.9\\
        \hline
        5 & 54 & 90.1 \\
        \hline
        6 & 20 & 91.3 \\
        \hline
        all & 166 & 88.0 \\
        \hline
    \end{tabular}
    \label{speedating}
\end{table}





\subsection{Applied problems}
\subsubsection{Deception study}

\begin{table*}[h]
  \centering
  \caption{Deception Study Witness Eye Gaze Region Frequencies During Relevant Questioning}
  \begin{tabular}{|c|c|c|c|c|c|c|}
    \hline
    Gaze Region & \makecell{Mean BLUFF} & \makecell{Mean TRUTH} & \makecell{STD BLUFF} & \makecell{STD TRUTH} &  T-test p-val & \makecell{Cohen's d\\ effect size}\\ [0.5ex]
    \hline
    region 1 & 0.001 & 0.001 & 0.004 & 0.001 & 0.193 & 0.286\\
    region 2 & 0.009 & 0.009 & 0.013 & 0.015 & 0.967 & 0.009\\
    region 3 & 0.001 & 0.002 & 0.003 & 0.003 & 0.771 & -0.064\\
    region 4 & 0.053 & 0.048 & 0.039 & 0.042 & 0.573 & 0.123\\
   	region 5 & 0.866 & 0.844 & 0.065 & 0.076 & 0.148 & 0.318\\
    region 6 & 0.041 & 0.039 & 0.043 & 0.043 & 0.821 & 0.05\\
    region 7 & 0.005 & 0.005 & 0.016 & 0.008 & 0.943 & -0.016\\
    \textbf{region 8} & \textbf{0.020} & \textbf{0.047} & \textbf{0.018} & \textbf{0.047} & \textbf{0.0004} & \textbf{-0.794}\\
    region 9 & 0.003 & 0.006 & 0.005 & 0.012 & 0.109 & -0.353 \\
    \hline
  \end{tabular}
  \label{table_static}
\end{table*}

Table \ref{table_static} shows the fractions of time a witness' gaze in the deception study falls in each of the gaze regions during the relevant questioning phase. As shown, there is one gaze region, region 8 - looking down, that shows significant average difference in the between the BLUFF and TRUTH groups, with truthful witnesses looking down more often (p=0.0004). It should be noted that due to Bonferroni multiple comparison correction \cite{cabin2000bonferroni}, the p-values should only be deemed significant if they are less than $\alpha=0.0056$ (i.e., confidence level/\# comparisons = 0.05/9), which it is for region 8. The effect size for region 8 is 0.79, which is a "large" effect size \cite{cohen1988statistical}. 
\begin{table}[h]
  \centering
  \caption{Deception Detection Classification Accuracies for Different Feature Sets}
  \begin{tabular}{@{}|c|c|c|c| @{}}
    \hline
    \makecell{Features} & \makecell{Model} & \makecell{Test set\\ accuracy} & \makecell{Test set\\ log-loss} \\
    \hline
    Emotion features & L1 & 0.533 & 0.730\\
    Emotion features & L2 & 0.554 & 0.717\\
    Action Units & L1 & 0.522 & 0.732\\
    Action Units & L2 & 0.495 & 0.719\\
    ICE Region frequencies (IRF) & L1 & 0.637 & 0.678\\
    ICE Region frequencies (IRF) & L2 & 0.643 & 0.703\\
    IRF + Emotion features & L1 & 0.589 & 0.700\\
    IRF + Emotion features & L2 & 0.660 & 0.677\\
    \hline
  \end{tabular}
  \label{table_accuracy}
\end{table}

Shown in Table \ref{table_accuracy} are the classification accuracy results when running a logistic regression classifier using the different feature sets as described in the methods section. As shown, using the emotion and action unit features alone resulted in test set accuracies only slightly above random chance (55.4\% and 52.2\% respectively.) Using the ICE region frequencies alone, the the classification accuracy is 64.3\%. The best performance is achieved by combining the ICE region frequencies and emotion features yielding 66.0\% accuracy when using L2 regularization logisitic regression model.



\subsubsection{Speed Dating}
\begin{table}[h]
  \centering
  \caption{Speed Dating Ratings Test Set Prediction Errors (MSE) with and without ICE RVE Feature}
  \begin{tabular}{@{}|c|c|c| @{}}
    \hline
    \makecell{Rating} & \makecell{AU only features} & \makecell{AU + ICE RVE features}\\
    \hline
    communication skill & 1.307 & 1.268\\
    eye contact & 1.756 & 1.717\\
\hline
  \end{tabular}
  \label{speed_dating_mse}
\end{table}

Here we present the results of trying to improve the accuracy of predicting speed dating performance by adding the ICE RVE as a feature. Shown in Table \ref{speed_dating_mse} are the mean squared errors associated with predicting human-rated scores for communication skill and eye contact using facial features alone, and using facial features combined with the ICE RVE fraction. Values shown are the average of 5-fold cross-validation. As shown, adding ICE RVE as a feature provides a small reduction in test set mean squared error (RMSE). Shown in Figure \ref{regweights}ab are the LASSO model weights ($\beta$) of the features used to predict the eye contact (a) and conversational skill (b) ratings. As shown eye contact weight Figure 7a, the ICE RVE feature is the second highest magnitude feature (behind AU12 - lip corner puller).

\begin{figure*}[!t]
  \centering
  \includegraphics[width=\linewidth]{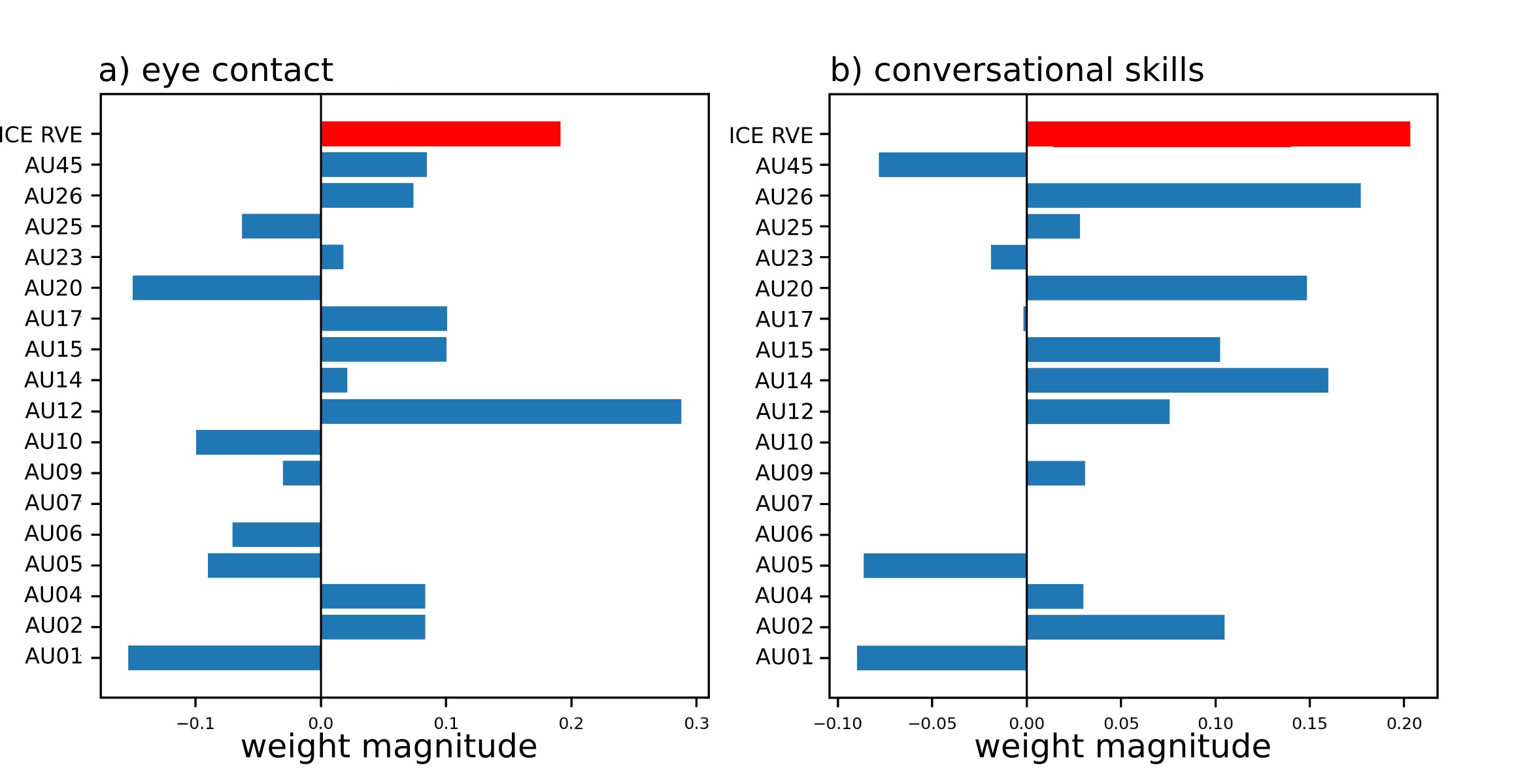}
  \setcounter{figure}{6}
  \caption{\textbf{Linear Regression (LASSO) Model $\beta$ magnitude comparison in prediction of expert human-rated a) eye contact rating, and b) conversational skills rating.} (Note that the ICE Eye\_con feature has the second largest magnitude for eye contact rating prediction and largest magnitude for conversational skills prediction.)}
  \label{regweights}
\end{figure*}

Figure \ref{regweights}b shows the weights of the features used in the LASSO model to predict the conversational skills ratings. From this figure we see that the eye contact feature has the highest weight among all the features. This indicates that the eye contact feature was more important than other facial expressions when applying a linear model to the speed dating scenario. 




\section{Discussion}
\subsection{The Region of Visual Engagement (RVE)}
The RVE is the densest area that an individual looking at during the conversation, which is the region of high interest for the individual. Assuming that a person looks at the conversation partner the most during the interaction, RVE is most likely to represent the conversation partner. However, there might be small variations of the RVEs between people since some only look at the eyes of the other while others might look at the entire face(s). In our video chat validation section, we define the RVE as the face of individuals on the screen, which is not necessarily correct in all cases. There should be some small variations in participants gaze behaviors that makes their RVE larger or smaller than the defined region (the face). This also shows the importance to validate the method with the human-annotated dataset, as human raters can better judge the variations of eye contact behaviors of individuals to give appropriate ratings.

On the other hand, since the RVE captures the region that an individual looks at the most during a conversation, it is intuitive that there should be a strong correlation between looking into the RVE and looking at the conversation partner's eyes. In fact, the area representing the eyes of the conversation partner should be a sub-region of RVE. In 5.3, we will demonstrate that RVE is an important signal that can facilitate the predictions of deception detection and conversational skill ratings.

\subsection{ICE Encoder Validation}

The results demonstrate that the ICE encoder is capable of successfully extracting interpersonal eye gaze from common video with neither specialized hardware nor prior knowledge of the physical layout. Not only do the validation results show that ICE works in video chat environments, but also in recordings of face-to-face interactions. While not directly comparable, for reference, the ICE IR-validation F1-score of 0.846 is on the order of the F1-score of specialized head-mounted hardware for detecting eye contact (see Ye, et al., with a calculated F1 score of 0.758) \cite{ye2012detecting}. The performance of ICE was reasonably robust given that the accuracy ranged between 61.1\% and 95.1\% percent agreement with the IR tracker ground truth (F1= 0.738 to 0.974), well above random chance. The standard deviation of the accuracy and F1 scores were 11.8\% and 0.0857 respectively, indicating that there is an expected 95\% chance that ICE will return an F1 score within 0.17 of the expected F1-score(0.846). 
In the speed-dating dataset, ICE shows an expected positive correlation of 0.37 between the estimated proportion of time people spent looking at their conversation partners and the eye contact ratings. The correlation of the mean average ICE percentages for each rating is 0.91.

It is important to note that the two validation tests (i.e. infrared eye gaze tracking and expert human-rated face-to-face videos) differ in two important ways. While the infrared eye gaze tracker provides an objective measure, the speed dating data set provides a subjective, human perspective. A human perspective is of important since even though the infrared eye tracker provides a high degree of accuracy, a raw analytic metric may not always be able to best gauge what is ultimately a human behavior, since slight deviations from normal may have large behavioral consequences. For example, interacting with someone who has a lazy eye, even though they may generally tend to make eye contact, the slight indirectness can cause substantial difference in human perceptions. While the infrared eye gaze tracker might not be tuned to understand which slight eye contact variances produce an uncanny valley type of effect, we would expect expert human raters to be able to do so. 

\begin{figure}[t]
  \centering
  \includegraphics[width=0.97\linewidth]{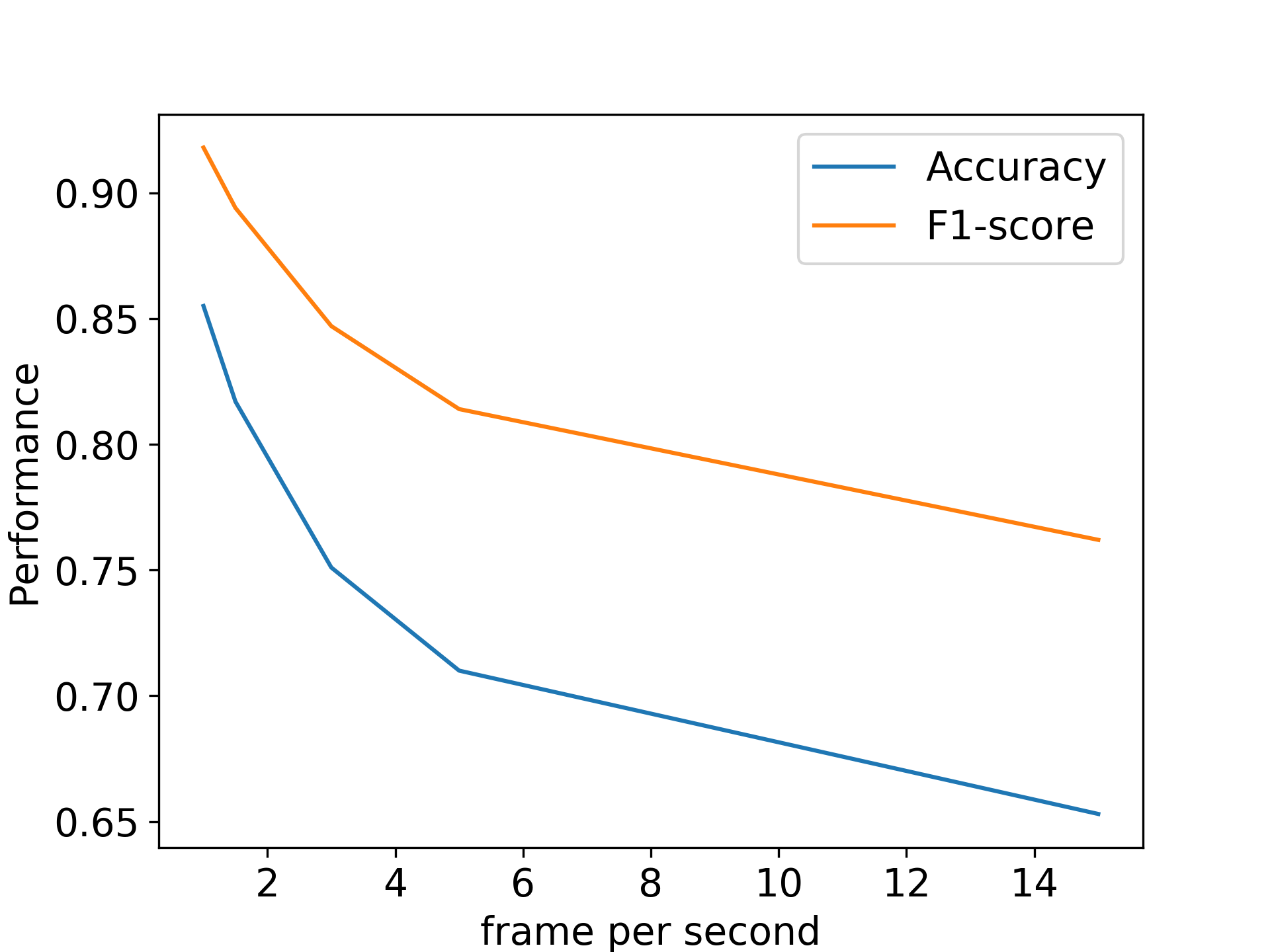}
  \caption{Frame-rate vs. Performance of ICE on per-frame basis}
  \label{fps}
\end{figure}

We downsample the data to minimize the effect of noises from OpenFace, by taking the majority votes of consecutive frames. Figure \ref{fps} shows the effect of frame-rate on the performance of ICE on the video chat validation data. However, if we downsample the data to a low frame rate, important information might be lost. We decide 3 fps is an appropriate frame rate as there are typically 2-4 eye fixations per seconds \cite{ford1959analysis}. Only video chat validation data are downsampled to compare to the ground-truth data on a frame by frame basis. Other experiments in the study use the mean \%RVE, which is not strongly affected by downsampling.

\begin{figure}[t]
  \centering
  \includegraphics[width=0.97\linewidth]{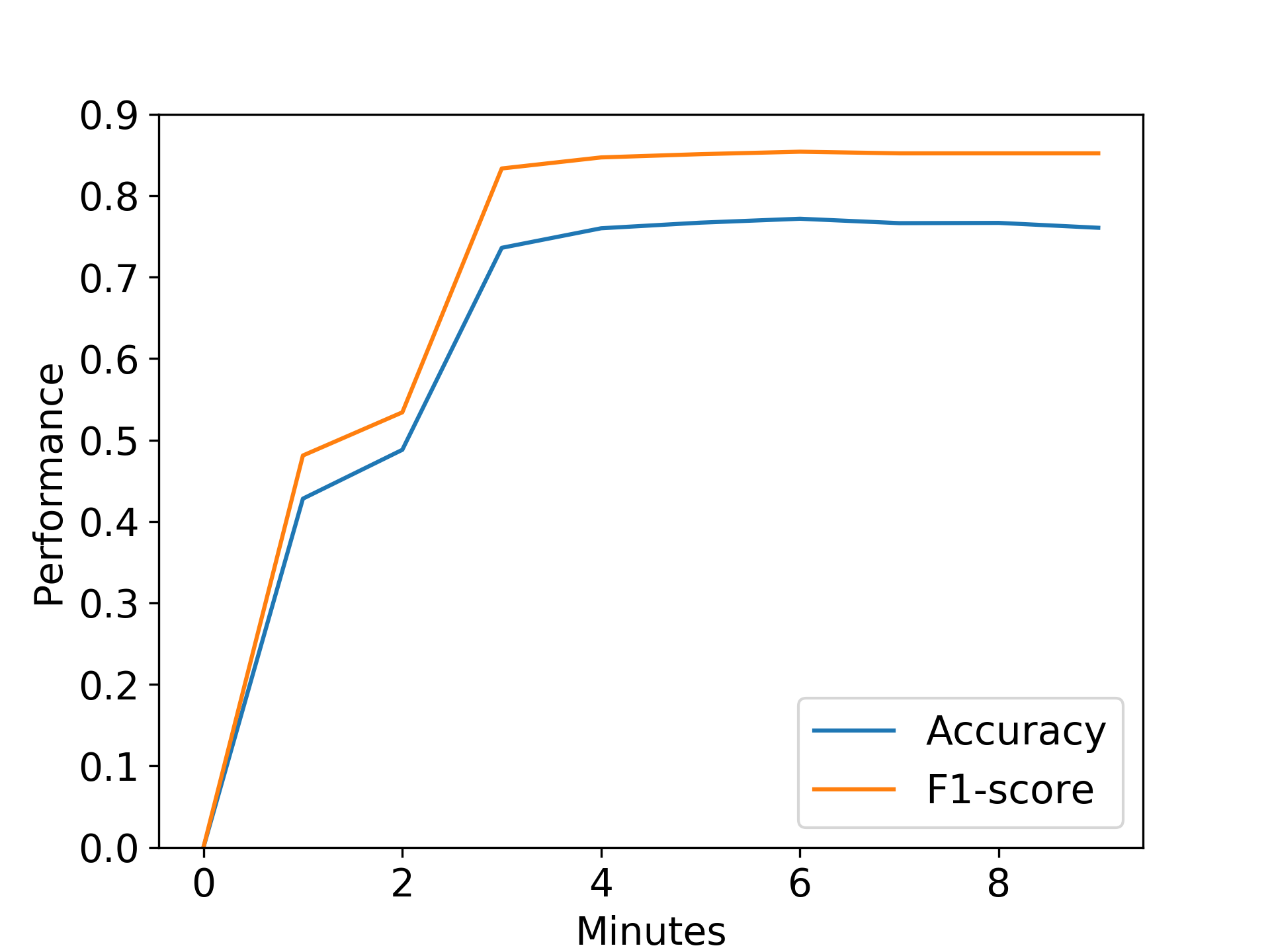}
  \caption{Performance of ICE when only using the first x minutes of recordings for training the encoder}
  \label{calibration}
\end{figure}

Figure \ref{calibration} shows the performance of the encoder on the 8 video chat validation recordings vs. the amount of time used two train the encoder. Specifically, we use the first x minutes of each recordings as input frames to the dyadic clustering algorithm, and report the performance of ICE on the whole recordings with the ground-truth provided by the infrared eye-tracker. It can be seen that the accuracy of the algorithm increases with the more time used to train, and the performance converge at around 3-4 minutes. For reference, Muller et al. report the calibration time of their framework to be around 6 minutes while Zhang et al. recommend to calibrate their method for approximately 3 hours.

\subsection{Applied problems}
The two applications that ICE was applied to occurred in two very different settings: a virtual, computer-mediated environment, and an in-person, physically close environment. By demonstrating that interpersonal gaze extraction in both of these environments revealed fresh insights, we highlight the utility of ICE as a gaze analysis tool in multiple domains.

\subsubsection{Deception Detection}
As discussed in the Background section, several studies have come to differing conclusions with regards to eye gaze in deceptoin, with some finding that eye contact is broken during deception \cite{ekman2009telling, zhang2015appearance, doherty2002development, doherty2005gaze}, while others have come to the conclusion that eye contact should increase in deception  \cite{kleinke1986gaze, sitton1981detection, bond1985miscommunication, granhag2002repeated}. Even though our results do not show a significant difference in the RVE (i.e. region 5, indicating one-way eye contact) between honest and and deceptive witnesses, we find that honest witnesses tend to look down more often than deceivers. Past research has established that eye movements play an important role in the visual recall memory process \cite{bulling2011recognition}. It has been shown that averting one's gaze during conversation is often done to facilitate memory recall since averting one's gaze helps to disengage someone from their surroundings (which can be distracting and impede memory recall) \cite{glenberg1998averting}. In the context of our experiment, these results make sense as truthful witnesses were given 60 seconds to memorize an image, and then were asked specific details from that image by the interrogator who asked scripted, memory-intensive questions. The following Figure \ref{lookdown} depicts a time slice of a truthful witness from our data answering the question, "If there was something to count in the image, what would it be and what would the count be?", after which the witness breaks eye contact and looks down, and after a pause begins to answer.

\begin{figure}[t]
  \centering
  \includegraphics[width=0.97\linewidth]{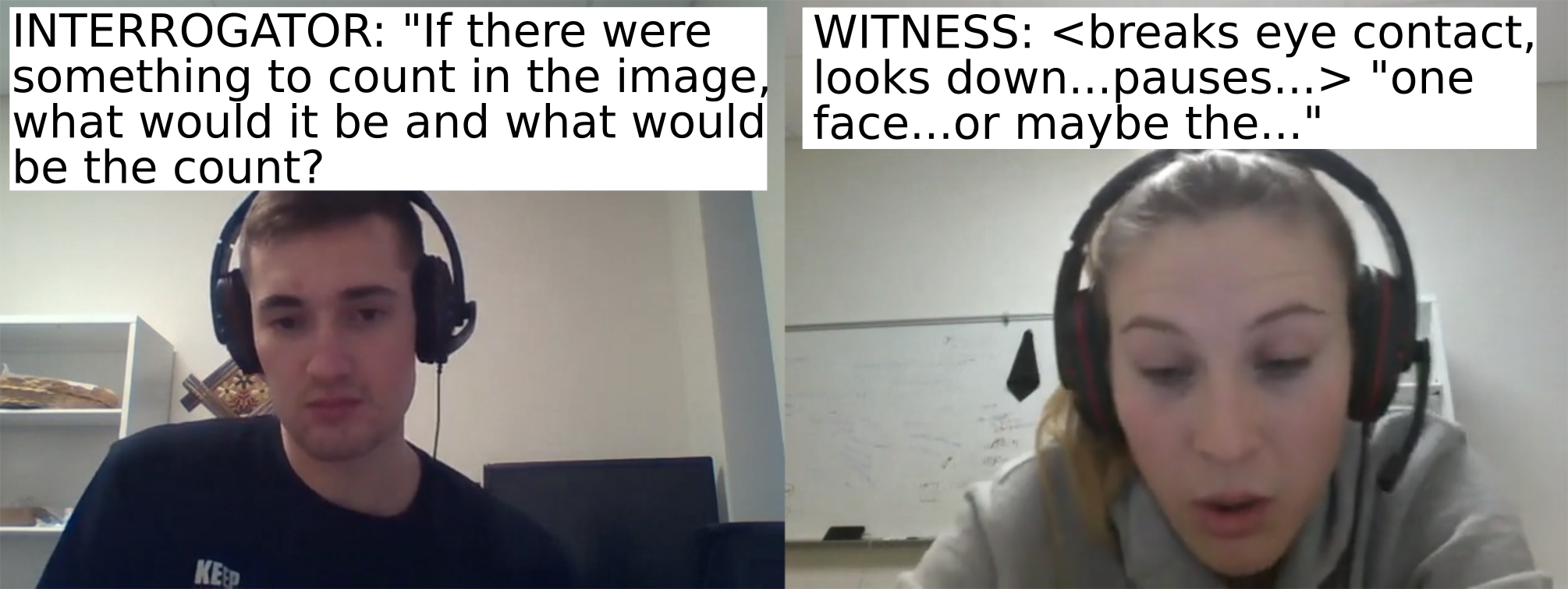}
  \caption{Truthful witness looking down as she recollects details of the evidence image}
  \label{lookdown}
\end{figure}

\subsubsection{Conversation Skills Prediction in Speed Dating}
The authors of the speed dating study \cite{ali2015lissa} demonstrated that expert human-rated communication skills ratings can be predicted from facial expressions. In our analysis, we also demonstrate that ICE encoded eye gaze can be combined with facial expressions to improve prediction. Further, out of all the features used (i.e. all facial expression action units and ICE RVE), ICE RVE has the highest model weight for predicting a speed daters conversational skills rating. This supports the common wisdom that making eye contact (or more specifically looking at the head region of your conversation partner) is an important aspect of conversational skills. Perhaps more importantly, this finding validates that the ICE framework is capable of generating meaningful gaze information in data in which the camera position is unknown.

\subsection{Limitations}
Even though this encoder shows utility as an analysis tool, there are some potential limitations. First, ICE should probably only be applied to dyadic video recordings in which the participants spend a significant portion of the time actively looking at each other. This is due to the fact that ICE relies upon their being a single dominant cluster of gaze area that will be detected as the region of primary visual engagement (RVE, i.e. region 5). Thus, the current configuration of ICE is limited to only two participants talking to each other and not group conversations. However, it should be noted that the encoder could be extended to detect a number of faces (N) by detecting the N largest clusters and in turn defining N regions of visual engagement. Second, ICE is not able to always identify a trusted region of visual engagement. As mentioned in the results, in 2.3\% of the speed dating videos, ICE was unable to identify a prominent cluster to represent gaze directed at a participant's conversation partner. 
Another limitation of the algorithm is its dependence on the ability to correctly extract eye-gaze signal from OpenFace (or a third-party software). However, there are several features of ICE that could reduce the effect of inconsistencies of eye-gaze extraction tools. We only work with frames that OpenFace reports a confidence of more than 0.9. Moreover, ICE uses DBSCAN for clustering, which is robust to various types of noises. Even if eye-gaze extraction tools are not totally reliable, they should at least be able to estimate the relative eye-gaze directions with respect to the camera position, which is sufficient for gaze clusters to form and be detected by the clustering algorithm.

\subsection{Contribution of ICE in Light of Related Work}
To our knowledge, ICE is the first framework to provide interpersonal gaze detection from video recordings in both video chat and in-person environments where the physical layout is not known a priori. In addition, ICE is the first system to provide interpersonal gaze which has been validated with the gold standard IR gaze tracking. Muller et al.'s work was instrumental in identifying that with eight cameras gaze at multiple conversation targets can be identified using automatic speaker labeling using facial expression analysis. However, it is unclear whether their method can be extended to video chat settings. Additionally, Muller et al.'s algorithm relies on each speaker speaking for a significant portion of time (in order to soft-label speaker segments for model training). In a dyadic setting, if only one speaker is speaking, the Muller algorithm will be unable identify when silent speakers are being looked at, and thus unable to detect interpersonal gaze. Indeed, in many conversational settings, floor time is often unbalanced. We believe Zhang et al. made a remarkable discovery that gaze clustering can be used to detect on-target gaze. However, their framework relies on the target being located close to the camera. ICE makes no requirement or assumptions about where the targets or cameras are located. Additionally, Zhang et al. recommend that three hours of each video setting be used for calibration/training their system. In our analysis, video clips are less than 15 minutes long and thus insufficient length for such a calibration. Indeed, Zhang et al. uses the OPTICS clustering algorithm to identify the "on target" cluster. In analyzing the ability to use Zhang et al's algorithm on our data sets we ran OPTICS on our raw gaze data and found that when running OPTICS (sklearn) on our data set videos, more than 80\% of videos ended up having all points (not just on-target points) in a single cluster. This indicates that in some settings, a framework which relies on OPTICS may not be able to distinguish interpersonal gaze from looking away.

It should also be noted that to our knowledge ICE is the first framework to provide \textit{relative} interpersonal gaze from standard video recordings, in that ICE provides whether gaze is above/below, to the left/right, as well as four diagonal directions, in addition to indicating whether the gaze is directed at the conversation partner (in contrast to prior work systems which merely provide Boolean indication of whether gaze is on a target or not). This point is particularly relevant, since our finding in the deception data set that honest witnesses tend to look down from their conversation partner more than bluffing witnesses would not have been identified if we relied on Boolean "on target" interpersonal gaze alone.

We compare our framework to the performance of the Zhang et al. and Mueller et al. references discussed. Zhang et al. evaluate their method on two datasets: object-mounted webcam with object targets and head-mounted webcam with human targets \cite{zhang2017everyday}. With human targets, their algorithm achieves MCC scores ranging from 0.1 to 0.3. For reference, our algorithm achieves an average MCC score of 0.23 on the 8 infrared eye gaze tracker validation videos. It should be noted that the experimental setup of our validation dataset (object-mounted webcam with human targets) is different from Zhang et al.'s dataset (head-mounted webcam with human targets). 

Mueller et al. alternatively measure their performance in accuracy and report an accuracy of approximately 0.7 in multi-person interaction scenario (4-person conversation in their recordings) \cite{muller2018robust}. For comparison, our framework's accuracy is 0.76 on the infrared-based validation videos, but it is currently applicable to only dyadic interactions.

\subsection{Future Work}
We envision that ICE may be applied to the domain of automated social skills development tools. In many social skills development tools having an interpersonal eye gaze tracking is very important \cite{ali2018aging}.  Lack of "proper" eye contact is associated with several behavioral conditions including social anxiety \cite{heimberg2002cognitive},  schizophrenia \cite{smith1996social}, and autism. In addition, ineffective nonverbal communication can impair positive social relationships  \cite{schwartz2017emotion, keating2016developmental, struchen2011examining}. 
Having a tool that can automatically give information about the close estimation of eye contact can potentially transform the social skills training at a large scale. Automated coaching systems have shown promise in developing communication skills in the domains of job interview \cite{hoque2013mach}, individuals with autism \cite{ali2015lissa}, and elderly individuals hoping to regain lost skills to help overcome social isolation \cite{ali2018aging}. Part of the benefit of these automated coaching systems is that it allows users to practice in their own environment (e.g. home), without the stigma of other people watching them. The ICE encoder is likely to be able to enhance such systems, by allowing interpersonal gaze to be accurately measured with commonly available webcam hardware, in environments for which the physical layout is unknown. ICE could be further developed as a passive application to monitor kids eye-gaze patterns as kids interact with computers/phones. If irregular and inconsistent eye gaze patterns are noticed, parents could be alerted (e.g., early signs of autism could include irregular eye gaze patterns).  

In addition to augmenting social skills development tools, we envision applying ICE to analyze group discussion environments. This would allow us to answer questions such as: \textit{To what extent does eye gaze affect group conversation metrics, such as productivity of the meeting, and participant attitudes towards each other?} Ideally this new knowledge would be applicable to facilitate improved group dynamics resulting in the generation of novel, impactful ideas more efficiently. Another exciting area for future work would be extending the capabilities of \textit{interpersonal computing systems} which aim to unobtrusively monitoring the emotional climate of a classroom  \cite{di2018unobtrusive, gashi2018using}. An extended version of ICE may be able to gauge whether an audience is engaged with a presentation\cite{gashi2019using}.

\section{conclusion}
In summary, we proposed a novel method of interpreting eye-gaze activity by interpersonally calibrating an eye-gaze signal and demonstrated the utility of this transformation by applying ICE to the in-person speed dating and computer-mediated deception detection domains. Our method automatically calibrates eye-gaze signals relative to a conversational partner in the form of a discrete encoder. We validate this encoder both objectively with an infrared eye gaze tracker in online video chats, as well as subjectively with expert human raters in an in-person, face-to-face setting. We show the value of ICE by applying the encoder to two different datasets each representing a different domain of human communication and physical set-up to understand the role of eye-gaze specifically in each setting. Our algorithmic contributions may serve as an initial step towards  broadening the utility of eye-gaze behaviors using our every day devices. The findings may motivate interpersonal gaze-calibration as a mechanism to explore interpersonal gaze relationships in multiple domains and contexts.

\ifCLASSOPTIONcaptionsoff
  \newpage
\fi



%


\bibliographystyle{IEEEtran}
\bibliography{refs.bib}

\begin{IEEEbiography}[{\includegraphics[width=1in,height=1.25in,clip,keepaspectratio]{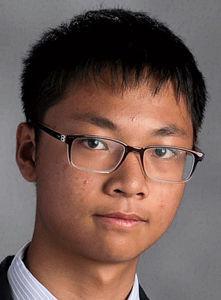}}]{Minh Tran}
Minh Tran is currently working towards a dual BS degree in Applied Mathematics and Computer Science at the University of Rochester. He is focusing his undergraduate research on machine learning models of nonverbal communication and natural language processing. Minh was the recipient of National Science Foundation undergraduate funding for the summer of 2019 as an undergraduate researcher at Duke University investigating the physics of X-ray diffraction using machine learning models.
\end{IEEEbiography}
\vspace{-15 mm}
\begin{IEEEbiography}[{\includegraphics[width=1in,height=1.25in,clip,keepaspectratio]{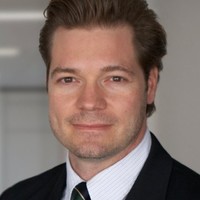}}]{Taylan Sen}
Taylan Sen received the BS degree in Electrical Engineering and MS degree in Biological Engineering from Cornell University and the JD degree from University at Buffalo. He has seven years industry experience as a software engineer and five years experience as an intellectual property attorney. He is currently working toward the PhD degree in Computer Science at University of Rochester. His research focuses on computational models of nonverbal communication and is conducted under the supervision of Prof. Mohammed Ehsan Hoque.
\end{IEEEbiography}
\vspace{-15 mm}
\begin{IEEEbiography}[{\includegraphics[width=1in,height=1.25in,clip,keepaspectratio]{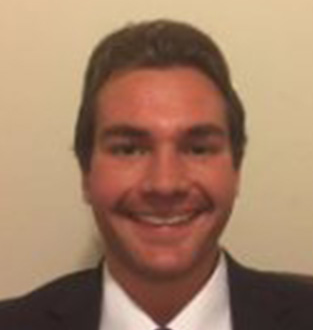}}]{Kurtis Haut}
Kurtis Haut received his B.A in Computer Science and Business from the University of Rochester in 2018. He is currently pursuing a joint PhD in Computer Science and Brain/Cognitive Science. His research interests include Computational Neuroscience, Human Cognition/Behavior, Artificial Intelligence, and Human Computer Interaction. 
\end{IEEEbiography}
\vspace{-15 mm}
\begin{IEEEbiography}[{\includegraphics[width=1in,height=1.25in,clip,keepaspectratio]{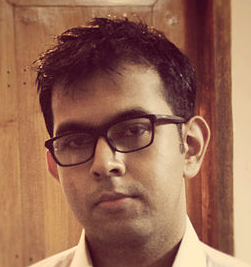}}]{Mohammad Rafayet Ali}
Mohammad Rafayet Ali received his B.Sc. in Computer Science and Engineering from Bangladesh University of Engineering and Technology in 2013 and MS degree from University of Rochester in 2016. He has worked as a Lecturer at Ahsanullah University of Science and Technology, Dhaka, Bangladesh. He is currently a PhD candidate at University of Rochester in the Computer Science Department. His research topics include AI approaches to understanding communication skills and the development of virtual agents for conversational skill coaching and evaluation. 
\end{IEEEbiography}
\vskip 0pt plus -1fil
\begin{IEEEbiography}[{\includegraphics[width=1in,height=1.25in,clip,keepaspectratio]{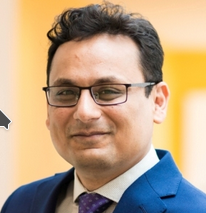}}]{Mohammed Ehsan Hoque}
Mohammed Ehsan Hoque received the Ph.D. degree from the Massachusetts Institute of Technology, in 2013. He is an assistant professor of computer science with the University of Rochester where he leads the ROC HCI Group. Hoque’s research is around developing computational tools to recognize the subtle nuances of human communication with a direct application of improving human ability.  His research has been recognized with the MIT TR35 Award, NSF CAREER Award, ECASE-Army Award He is a member of the ACM, IEEE and AAAI.
\end{IEEEbiography}






\end{document}